\documentclass[10pt,twocolumn,letterpaper]{article}

\usepackage[pagenumbers]{cvpr} % To force page numbers, e.g. for an arXiv version

\usepackage{multirow}
\usepackage{pifont}
\usepackage[normalem]{ulem}
\useunder{\uline}{\ul}{}

\definecolor{cvprblue}{rgb}{0.21,0.49,0.74}
\usepackage[pagebackref,breaklinks,colorlinks,allcolors=cvprblue]{hyperref}

\def\paperID{1870} % *** Enter the Paper ID here
\def\confName{CVPR}
\def\confYear{2025}

\title{UniBrain: A Unified Model for Cross-Subject Brain Decoding}
\author{Zicheng Wang\textsuperscript{\rm 1}~~~~Zhen Zhao\textsuperscript{\rm 1}~~~~Luping Zhou\textsuperscript{\rm 2}~~~~Parashkev Nachev\textsuperscript{\rm 3} \vspace{2mm} \\ 
\textsuperscript{\rm 1}Shanghai Artificial Intelligence Laboratory \\
\textsuperscript{\rm 2}University of Sydney\hspace{16mm}
\textsuperscript{\rm 3}University College London\\
{\tt\small \{xiaoyao3302, zhen.zhao\}@outlook.com~~luping.zhou@sydney.edu.au~~p.nachev@ucl.ac.uk}
}
\begin{document}
\maketitle
\begin{abstract}
Brain decoding aims to reconstruct original stimuli from fMRI signals, providing insights into interpreting mental content. 
Current approaches rely heavily on subject-specific models due to the complex brain processing mechanisms and the variations in fMRI signals across individuals.
Therefore, these methods greatly limit the generalization of models and fail to capture cross-subject commonalities. 
To address this, we present UniBrain, a unified brain decoding model that requires no subject-specific parameters. 
Our approach includes a group-based extractor to handle variable fMRI signal lengths, a mutual assistance embedder to capture cross-subject commonalities, and a bilevel feature alignment scheme for extracting subject-invariant features. 
We validate our UniBrain on the brain decoding benchmark, achieving comparable performance to current state-of-the-art subject-specific models with extremely fewer parameters. 
We also propose a generalization benchmark to encourage the community to emphasize cross-subject commonalities for more general brain decoding.
Our code is available at \url{https://github.com/xiaoyao3302/UniBrain}.
\end{abstract}    
\section{Introduction}
\label{sec:intro}

Brain decoding of visual perception, which aims to reconstruct visual stimuli from functional magnetic resonance imaging (fMRI) signals, has garnered significant attention due to its potential to reveal insights into human visual processing. 
This area of research bridges neuroscience and vision, offering a unique perspective on how complex real-world environments are represented in the brain.  
It holds promises for practical applications like brain-computer interfaces and vision-assistive technologies, demonstrating substantial scientific and practical utility in medicine and industry~\cite{kay2008identifying, haynes2006decoding, naselaris2011encoding, kamitani2005decoding}. 
Recent advancements in diffusion models, which excel in high-fidelity image generation, have inspired new methods for extracting latent embeddings from fMRI signals to guide these models in reconstructing visual stimuli with promising performance~\cite{takagi2023high, luo2024brain, huo2024neuropictor}.

\begin{figure}[t]
\centering
\includegraphics[width=0.9\linewidth]{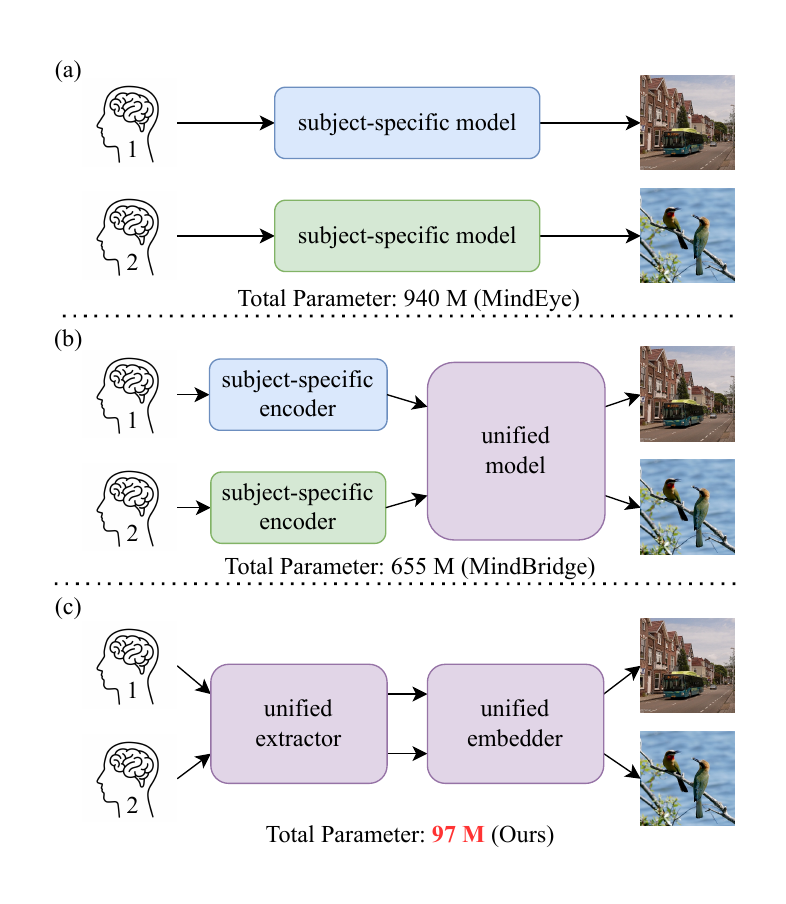}
\caption{Current brain decoding methods either train separate models for each subject (a) or train a partially unified model with subject-specific modules (b). Such methods greatly limit the generalization of the model with little scalability and struggle to capture cross-subject commonalities. In this paper, we for the first time propose a unified model for fMRI decoding cross subjects which requires \textbf{no subject-specific modules} (c), which significantly reduces the number of parameters while capturing the potential pattern across different subjects, and can be directly adopted for brain decoding on new subjects.} 
\label{fig_intro_1}
\end{figure}

The fMRI signals vary significantly across individuals due to genetic and environmental influences on brain systems~\cite{mueller2013individual}, as well as differences in signal length across subjects~\cite{allen2022massive, finn2017can}.
These issues pose significant challenges to designing a unified brain decoding model.
Therefore, current brain decoding methods can be broadly divided into two categories.
The first category includes models like MindEye (Fig.~\ref{fig_intro_1} (a)) that train separate models for each individual, effectively capturing subject-specific patterns but struggling to generalize across individuals, making them resource-intensive and less efficient~\cite{scotti2024reconstructing, ozcelik2023natural, lu2023minddiffuser}. The second category, such as MindBridge (Fig.~\ref{fig_intro_1} (b)), introduces partially unified models with subject-specific modules~\cite{wang2024mindbridge, xia2025umbrae, scotti2024mindeye2}. However, relying on subject-specific parameters, these methods only partially address cross-subject generalization challenges. They lack scalability and cannot easily extend to new subjects not seen during training, limiting their ability to capture cross-subject commonalities in brain function~\cite{geerligs2015state, ruffle2024computational}.

Facing the issue, in this paper, we propose a unified model for cross-subject fMRI decoding (UniBrain), \textbf{eliminating the need for subject-specific parameters} for the first time.
As shown in Fig.~\ref{fig_intro_1} (c), our method consists of a unified extractor to extract latent representations across subjects and a unified embedder to convert fMRI representations into embeddings to guide diffusion model-based stimulus reconstruction.
Our unified model efficiently captures shared patterns across individuals, significantly reduces model parameters, and generalizes directly to new subjects.
However, building a unified model to capture cross-subject commonalities poses challenges. 
First, the varying length of fMRI signals necessitates a model that can handle variable-sized inputs.
Second, there is no explicit guarantee that features extracted from fMRI signals are inherently subject-invariant.
To address \textbf{the first challenge}, leveraging the fact that neighbouring voxels in fMRI signals share similar functional selectivity~\cite{ccelik2019spatially}, we introduce a voxel aggregation operation to aggregate fMRI voxels into a fixed number of groups, standardizing signal length across subjects.
This allows fMRI signals to be analyzed by our newly proposed unified group-based extractor to obtain latent representations.
To tackle \textbf{the second challenge}, we propose a bilevel feature alignment scheme: at the extractor level, we use adversarial training to make representations indistinguishable by a subject discriminator, while at the embedder level, we map embeddings to a common feature space, \textit{i.e.}, the CLIP space~\cite{radford2021learning}, ensuring subject-invariant feature extraction.

Nevertheless, the low spatial resolution of fMRI signals~\cite{kamitani2005decoding, haynes2005predicting} complicates direct alignment with CLIP image features, often resulting in suboptimal guidance for the reconstruction process.
Semantic and geometric information, however, can complement each other — semantic content provides high-level context, while geometric details add fine-grained precision~\cite{horikawa2017generic}.
Inspired by this, we introduce a Transformer-based unified mutual assistance embedder to decode fMRI representations into both semantic and geometric embeddings in a coarse-to-fine manner, with each type of information assisting the other.
Initially, the fMRI signals are separately decoded into coarse semantic and geometric embeddings. 
Then, a mutual embedder aggregates the information from both, producing fine semantic and geometric embeddings.
The embeddings are aligned with CLIP text and image features, respectively, enabling the embeddings to provide enriched precise guidance for stimulus reconstruction.

We validate the effectiveness of our UniBrain using the widely used brain decoding NSD benchmark~\cite{allen2022massive}.
Under the common in-distribution setting with seen subjects, our UniBrain achieves comparable decoding performance to current state-of-the-art (SOTA) methods, despite using less than 20\% of their parameters and requiring no subject-specific modules.
Furthermore, we develop the cross-subject out-of-distribution (OOD) setting, a new generalization benchmark to assess the model's ability to generalize on unseen subjects. 
We aim to encourage the community to place greater emphasis on cross-subject commonalities and their significance for advancing generalized brain decoding methodologies.
We hope such studies could contribute to more practical utility in medicine and industry.

The contributions of our work are summarized below:
    \begin{itemize}
    \item We introduce a novel unified model for cross-subject brain decoding (UniBrain) that eliminates the reliance on subject-specific parameters, firstly enabling effective cross-subject out-of-distribution (OOD) decoding.
    \item Our model, requiring significantly fewer parameters, achieves comparable performance to state-of-the-art subject-specific methods in the common in-distribution (seen subject) setting.
    \item We develop a new generalization benchmark for brain decoding, advocating for greater emphasis on cross-subject commonalities in more general brain decoding research.
\end{itemize}

\section{Related Works}
\label{sec:review}

The contents of perception are believed to be encoded in the brain~\cite{shen2019deep}.
The research on brain decoding targets finding a reverse process to enable us to understand how much information the brain perceives from the world, which has attracted significant attention over the past decades~\cite{naselaris2011encoding, shirakawa2024spurious}.
Most studies focus on functional magnetic resonance imaging (fMRI) decoding due to its ability to measure brain activities through blood-oxygen-level-dependent (BOLD) signals~\cite{engel1994fmri, amaro2006study, naselaris2011encoding, shen2019end}. 

\begin{figure*}[t]
\centering
\includegraphics[width=0.9\linewidth]{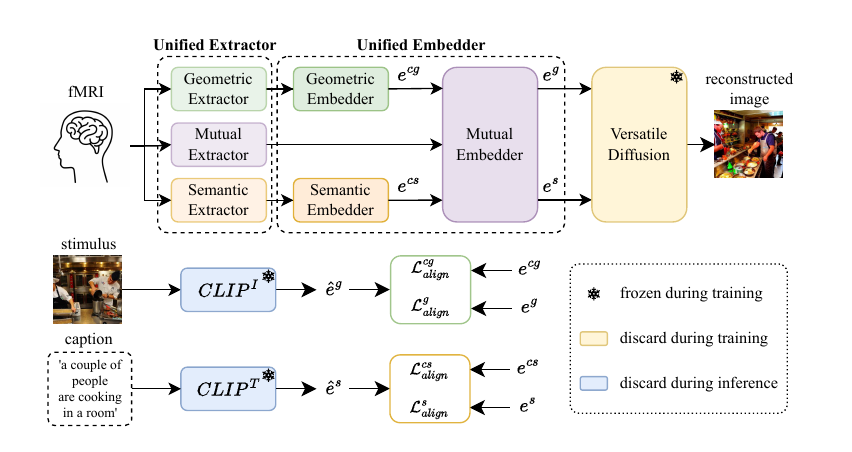}
\caption{An overview of our proposed unified model for cross-subject brain decoding (UniBrain), which includes a unified extractor and a unified mutual assistance embedder. The unified extractor includes three group-based extractors to extract semantic, geometric and mutual global representations from fMRI signals, respectively. The unified mutual assistance embedder includes two Transformer-based embedders to decode coarse semantic and geometric embeddings, respectively, and one Transformer-based mutual embedder to decode fine semantic and geometric embeddings. The mutual embedder takes as input the mutual global representation, the coarse semantic and geometric embeddings. The decoded embeddings can be used as the guidance of the Versatile Diffusion model for stimulus reconstruction. Note that the CLIP models only provide supervision during the training process, which are not utilized during the inference stage. Besides, the weights of the CLIP models and the Versatile Diffusion model are frozen.}
\label{fig_pipeline}
\end{figure*}

Early research decodes the fMRI signals into the shapes, the categories or the orientations of the objects~\cite{kamitani2005decoding, miyawaki2008visual}. However, the decoding outputs are confined to the classes of the decoder, limiting the scalability of the decoder to new unseen categories~\cite{miyawaki2008visual, horikawa2017generic, kay2009can}. Therefore, an image-level decoding model is desired.

Various previous studies have proved the feasibility of image-level decoding~\cite{nishimoto2011reconstructing, lin2022mind, beliy2019voxels}. Most of these methods follow a standard encoder-decoder framework.
They first extract latent features from the fMRI signals and then reconstruct the original stimulus with a generator~\cite{shirakawa2024spurious}. 
Currently, the success of Diffusion models~\cite{ho2020denoising, rombach2022high, peebles2023scalable} to generate high-quality images provides a powerful tool for brain decoding, inspiring tremendous research to utilize Diffusion models for stimulus reconstruction~\cite{zeng2024controllable, fang2024alleviating, chen2024mind, lu2023minddiffuser}. 
These methods extract latent representations from fMRI signals and use the representations as guidance for the Diffusion models to reconstruct the original stimuli with high fidelity. 

As is known, the brains of different individuals vary significantly, including both the length of the collected fMRI signals and the processing mechanism~\cite{allen2022massive, finn2017can}.
Therefore, current methods either train separate models for different individuals or train a partially unified model with subject-specific parameters. 
Such methods greatly limit the generalization of the model with little scalability, indicating the trained model can hardly be adopted on an unseen subject. Besides, such models will only examine the latent patterns of the examined patients, while cannot reflect the cross-subject commonalities~\cite{ruffle2024computational}. Therefore, a unified model that captures patterns across subjects is desired.

However, whether it is possible to train a brain decoding model that can capture population-level commonality and can directly be adopted by new individuals is still an open issue~\cite{ruffle2024computational}. Therefore, we come up with a generalization benchmark for brain decoding, encouraging the community to pay more attention to the generalized brain decoding research.
We emphasise the population-level commonalities and differences in human brains. Such research may contribute to more practical utility in medicine and industry.
\section{Methodology}
\label{sec:method}

% In this section, we will first give a brief introduction to the task of brain decoding in Sec.~\ref{sec:method_intro}. Then, we will introduce our newly proposed unified model for cross-subject brain decoding (UniBrain) in detail, which includes a group-based extractor in Sec.~\ref{sec:method_extractor}, a mutual assistance embedder in Sec.~\ref{sec:method_embedder}, and a bilevel feature alignment scheme in Sec.~\ref{sec:method_DG}.

\subsection{Problem Statement}
\label{sec:method_intro}
In brain decoding, we are given a set of fMRI signals $\mathcal{X} = \left\{ x_{i, j} \right\}_{i=1, j=1}^{N, M_i}$ and the corresponding stimuli images $\mathcal{Y} = \left\{ y_{i, j} \right\}_{i=1, j=1}^{N, M_i}$ collected from subject $i$ facing the stimulus $j$, $N$ indicates the number of subjects and $M_i$ indicates the number of stimuli for subject $i$. The fMRI signal has been preprocessed to a 1-D tensor, \textit{i.e.}, $x_{i, j} \in \mathbb{R}^{1\times D_i}$, with varying lengths across different subjects. Our goal is to train a model capable of reconstructing the original stimuli images using the given fMRI signals from different subjects. 

\subsection{Group-Based Extractor}
\label{sec:method_extractor}
Due to genetic and environmental factors, the brain's processing mechanisms differ among individuals. 
Besides, the length of fMRI signals across subjects also shows significant variations. 
Therefore, it is extremely difficult to design a unified model to capture cross-subject representations. 
Previous methods either used the pooling method to map the fMRI signals of different lengths to a fixed length~\cite{wang2024mindbridge}, or trained different mapping layers for each subject~\cite{xia2025umbrae}. However, the previous methods would lead to a loss of information, and the latter methods would limit the generalization of the model. Therefore, in this paper, we propose a new unified group-based extractor to capture cross-subject representations with minimum information loss.

Specifically, considering that the activities of neighbouring voxels in fMRI signals are similar~\cite{ccelik2019spatially}, we first uniformly select a set of key voxels from different fMRI signals with a fixed number of $G$ across subjects. 
Then, we calculate the distance of each voxel to the key voxel and aggregate $K$ voxels within a neighbouring area of a certain key voxel according to the calculated distance. 
Aggregating these voxels as a group will lead to a grouped fMRI signal with a shape of $1 \times G \times K$, indicating each fMRI signal is divided into $G$ groups and each group consists of $K$ voxels. 
The voxels within a group may exhibit a similar representation, and each group may represent a certain pattern. 
Thereafter, we design a group-based local extractor to extract latent local representations for each group.
We map the grouped fMRI signals to a latent space with a shape of $1 \times G \times D_l$, where $D_l$ indicates the dimension of the local representations of each group.

In addition, considering that a valid representation of brain activity requires various neurons~\cite{vinje2000sparse}, we aggregate the local patterns of each group to obtain the global representation of the brain activity.
In particular, we first flatten the local representations to a shape of $1 \times (G \times D_l)$, and then map the flattened local representations to the global brain representation $\bar{x}_{i, j}$ using a global brain extractor. Note that $\bar{x}_{i, j} \in \mathbb{R}^{1 \times D_b}$, $D_b$ indicates the dimension of the global feature of the fMRI signals. 
For the simplicity of introduction, we call the group-based extractor the combination of the group-based local extractor and the global brain extractor, denoted by $\psi$. 
Thereby, the extracted latent representations of the fMRI signals can be presented as $\bar{x}_{i, j} = \psi (x_{i, j})$.

\subsection{Mutual Assistance Embedder}
\label{sec:method_embedder}

After extracting the global brain representation $\bar{x}_{i, j}$ from the fMRI signal, we further focus on converting the fMRI representations into embeddings to guide the stimulus reconstruction. Most of the previous methods extract embeddings from the fMRI representations and align the embeddings with the corresponding CLIP image features~\cite{wang2024mindbridge, scotti2024reconstructing}, while the spatial resolution of fMRI signals is not high enough to decode fine-grained images~\cite{kamitani2005decoding, haynes2005predicting}, leading to difficulties on aligning the decoded embeddings with the corresponding CLIP image features.

In this paper, considering that semantic information can assist geometric information extraction by providing high-level semantic context and geometric information can also enhance semantic extraction by contributing fine-grained image-level details~\cite{horikawa2017generic}, we propose a new Transformer-based unified mutual assistance embedder that decodes fMRI representations into both semantic and geometric embeddings in a coarse-to-fine manner. 

Specifically, we first use a geometric extractor $\psi^g$ to project the fMRI signals to geometric representations $\bar{x}_{i, j}^g$, and then we use a geometric projector to project the geometric representations to geometric tokens with a shape of $1 \times T \times D_b$, $T$ indicates the number of tokens we map.
Next, we encourage our geometric embedder $\varphi^g$, which utilizes a cross-attention module~\cite{vaswani2017attention, alayrac2022flamingo} to decode coarse latent geometric embeddings $e_{i, j}^{cg}$, where $e_{i, j}^{cg} \in \mathbb{R}^{1 \times T_g \times D_b}$ and $T_g$ indicates the number of geometric embeddings. In particular, we project the geometric tokens as values $V^g$ and keys $K^g$, and also randomly initialize a set of learnable tokens as query $Q^g$. We encourage the updated queries as the output embeddings.
Similarly, we can also obtain the decoded coarse semantic embeddings $e_{i, j}^{cs}$ using a semantic embedder $\varphi_s$ where $e_{i, j}^{cs} \in \mathbb{R}^{1 \times T_s \times D_b}$ and $T_s$ indicates the number of semantic embeddings.

In addition, we target decoding a set of fine geometric embeddings and semantic embeddings based on the decoded coarse geometric embeddings and semantic embeddings. 
Therefore, we first project the fMRI signal to a set of mutual tokens using a mutual extractor $\psi^m$ and a mutual projector, then we concatenate the mutual tokens with the decoded coarse geometric and semantic embeddings as the input tokens to a mutual embedder $\varphi_m$, with an output of mutual embeddings $e_{i, j}^m$, where $e_{i, j}^{m} \in \mathbb{R}^{1 \times (T_g + T_s) \times D_b}$. Finally, we can divide the fine geometric embeddings $e_{i, j}^{g}$ and semantic embeddings $e_{i, j}^{s}$ from the mutual embeddings, which are then fed into the Versatile Diffusion model~\cite{xu2023versatile} to guide the stimuli reconstruction.

\subsection{Bilevel Feature Alignment}
\label{sec:method_DG}
Recall that we target extracting subject-invariant features to capture cross-subject commonalities. We thereby introduce a bilevel feature alignment scheme for subject-invariant feature extraction, including a subject-invariant representation extraction at the extractor level and a subject-invariant embedding extraction at the embedder level.

Considering that the stimuli of different individuals are different, we first perform a distribution-level feature alignment at the extractor level to extract subject-invariant global representations $\bar{x}_{i, j}$. 
In particular, we introduce a subject discriminator $\mathcal{D}$ to distinguish which subject the extracted representations belong to.  At the same time, we encourage the extractor $\psi$ to extract features that cannot be distinguished by the subject discriminator following an adversarial training manner. 
Recall that the subject label for each global representation is $i$ and $\bar{x}_{i, j} = \psi(x_{i, j})$, and the subject prediction of the subject discriminator can be presented by $\mathcal{D} (\psi(x_{i, j}))$. Such adversarial loss can be represented as:
\begin{equation}
    \mathcal{L}_{adv} = \frac{1}{N} \sum_{i=1}^{N} \frac{1}{M_i} \sum_{m=1}^{M_i}  \ell^{ce} (\mathcal{D} (\psi(x_{i, j})), i),
\end{equation}
where $\ell^{ce}$ indicates the cross-entropy loss.
Note that here we omit the superscript $s$, $g$ and $u$ for simplicity, and we adopt a gradient reversal layer~\cite{ganin2016domain} to perform adversarial optimization.

Besides, we also encourage a subject-invariant embedding extraction process to ensure that the embedders can capture cross-subject commonalities.
Recall that we have decoded four sets of embeddings, \textit{i.e.}, coarse geometric embeddings $e^{cg}$, coarse semantic embeddings $e^{cs}$, fine geometric embeddings $e^{g}$ and fine semantic embeddings $e^{s}$, here we encourage the embeddings to be aligned in a same feature space. 
Note that the stimuli of the NSD dataset are collected from the MS-COCO dataset~\cite{lin2014microsoft}, the textual captions $\mathcal{T} = \left\{ t_{i, j} \right\}_{i=1, j=1}^{N, M_i}$ are provided for each stimulus. 
At the same time, CLIP provides a large pre-trained vision-language model that effectively captures both visual and textual features, guaranteeing that features extracted from different images or texts lie in the same feature space.
Therefore, we encourage the embeddings extracted from different subjects to align with their corresponding CLIP features in the CLIP feature space.
Specifically, given the stimulus $x_{i, j}$ and the textual description $t_{i, j}$, we first adopt the pre-trained CLIP model to extract the corresponding visual and textual features $\hat{e}_{i, j}^g$ and $\hat{e}_{i, j}^s$, respectively. 
Then, we align the extracted coarse and fine geometric embeddings with the CLIP visual features, and also align the coarse and fine semantic embeddings with the CLIP textual features.
Such an operation encourages a subject-invariant embedding extraction. 
Following the previous works~\cite{xia2025umbrae, wang2024mindbridge, scotti2024reconstructing}, we use the combination of an MSE loss and a SoftCLIP loss to supervise the feature alignment. The MSE loss can be represented by:
\begin{equation}
    \mathcal{L}_{MSE}(\hat{e}^o, e^o) = \frac{1}{N} \sum_{i=1}^{N} \frac{1}{M_i} \sum_{m=1}^{M_i} \left(\hat{e}_{i, j}^o - e_{i, j}^o \right)^2,
\end{equation}
and the SoftCLIP loss can be represented by:
\begin{gather}
    \mathcal{L}_{SoftCLIP}(\hat{e}^o, e^o) = - \frac{1}{H} \sum_{\rho=1}^{H} \frac{1}{H} \sum_{\varrho=1}^{H} \notag \\ 
    \left[\frac{\mathrm{exp}\left(\frac{\hat{e}_{\rho} \cdot \hat{e}_{\varrho}}{\tau}\right)}{\sum_{\sigma=1}^{H} \mathrm{exp}\left(\frac{\hat{e}_{\rho} \cdot \hat{e}_{\sigma}}{\tau}\right)} \cdot \mathrm{log} \frac{\mathrm{exp}\left(\frac{e_{\rho} \cdot \hat{e}_{\varrho}}{\tau}\right)}{\sum_{\sigma=1}^{H} \mathrm{exp}\left(\frac{e_{\rho} \cdot \hat{e}_{\sigma}}{\tau}\right)}\right],
\end{gather}
where $\tau$ is a temperature hyperparameter. Here we treat each embedding as an entirety and use $H$ to replace $\sum_{i=1}^{N} M_i$, indicating the total number of embeddings. We also use the subscript $\rho$, $\varrho$ and $\sigma$ to replace $\left\{ i, j \right\}$ for simplicity, indicating different embeddings. Note that $o \in \left \{ cg, cs, g, s \right \}$ for extracted embeddings, indicating the coarse or fine geometric or semantic embeddings, respectively. Also note that we align both the coarse and fine embeddings with the corresponding CLIP embeddings, thus $o \in \left \{g, s \right \}$ for CLIP embeddings. We use $o \in \left \{ cg, cs, g, s \right \}$ for simplicity. All of the embeddings are L2-normalized. We can calculate the embedder-level alignment loss as:
\begin{equation}
    \mathcal{L}_{align}^{o} = \lambda_1 \mathcal{L}_{MSE}(\hat{e}^o, e^o) + \lambda_2 \mathcal{L}_{SoftCLIP}(\hat{e}^o, e^o),
\end{equation}
and
\begin{equation}
    \mathcal{L}_{align} = \sum_{o \in \left \{ cg, cs, g, s \right \}} \mathcal{L}_{align}^{o}.
\end{equation}
Incorporating the extractor-level adversarial loss, we calculate the total loss as
\begin{equation}
    \mathcal{L}_{total} = \lambda_0 \mathcal{L}_{adv} + \mathcal{L}_{align},
\end{equation}
to supervise our UniBrain model to capture subject-invariant features.

\begin{table*}[t]\centering
\scalebox{0.825}{
\begin{tabular}{lcccccccccc}
\toprule
\multirow{2}{*}{Method} & \multirow{2}{*}{Unified} & \multirow{2}{*}{Subj-agnostic} & \multicolumn{4}{c}{Low-Level} & \multicolumn{4}{c}{High-Level} \\ 
\cline{4-11} & & & PixCorr $\uparrow$ & SSIM $\uparrow$ & Alex(2) $\uparrow$ & Alex(5) $\uparrow$ & Incep $\uparrow$ & CLIP $\uparrow$ & EffNet-B $\downarrow$ & SwAV $\downarrow$ \\ 
\hline
Takagi et al.~\cite{takagi2023high} & \ding{55} & \ding{55} & - & - & 83.0\% & 83.0\% & 76.0\% & 77.0\% & - & - \\
Brain-Diffuser~\cite{ozcelik2023natural} & \ding{55} & \ding{55} & .254 & \textbf{.356} & 94.2\% & 96.2\% & 87.2\% & 91.5\% & .775 & .423 \\
MindEye~\cite{scotti2024reconstructing} & \ding{55} & \ding{55} & \textbf{.309} & {\ul.323} & \textbf{94.7\%} & \textbf{97.8\%} & {\ul 93.8\%} & 94.1\% & \textbf{.645} & \textbf{.367} \\
MindBridge~\cite{wang2024mindbridge} & \ding{51} & \ding{55} & .151 & .263 & 87.7\% & 95.5\% & 92.4\% & {\ul 94.7\%} & .712 & .418 \\
\hline
UniBrain$^\dag$ & \ding{51} & \ding{55} & \textbf{.309} & .317 & {\ul 94.2\%} & {\ul 97.4\% } & \textbf{94.5\%} & \textbf{95.3\%} & {\ul .656} & {\ul .374} \\
UniBrain (Ours) & \ding{51} & \ding{51} & .155 & .259 & 87.8\% & 95.5\% & 92.4\% & 94.0\% & .691 & .407 \\
\bottomrule
\end{tabular}
}
\caption{Quantitative comparison between our UniBrain and other brain decoding methods. All methods are trained and tested using four subjects. Metrics are calculated as the average across four subjects. ``Unified'' indicates whether the method utilizes a unified model or not. ``Subj-agnostic'' indicates whether the model requires no subject-specific parameters or not. The best performance is highlighted in bold and the second best result is underlined. $^\dag$ indicates our method using subject-specific extractors.}
\label{table_fully_supervise}
\end{table*}

\begin{figure*}[t]
\centering
\includegraphics[width=0.9\linewidth]{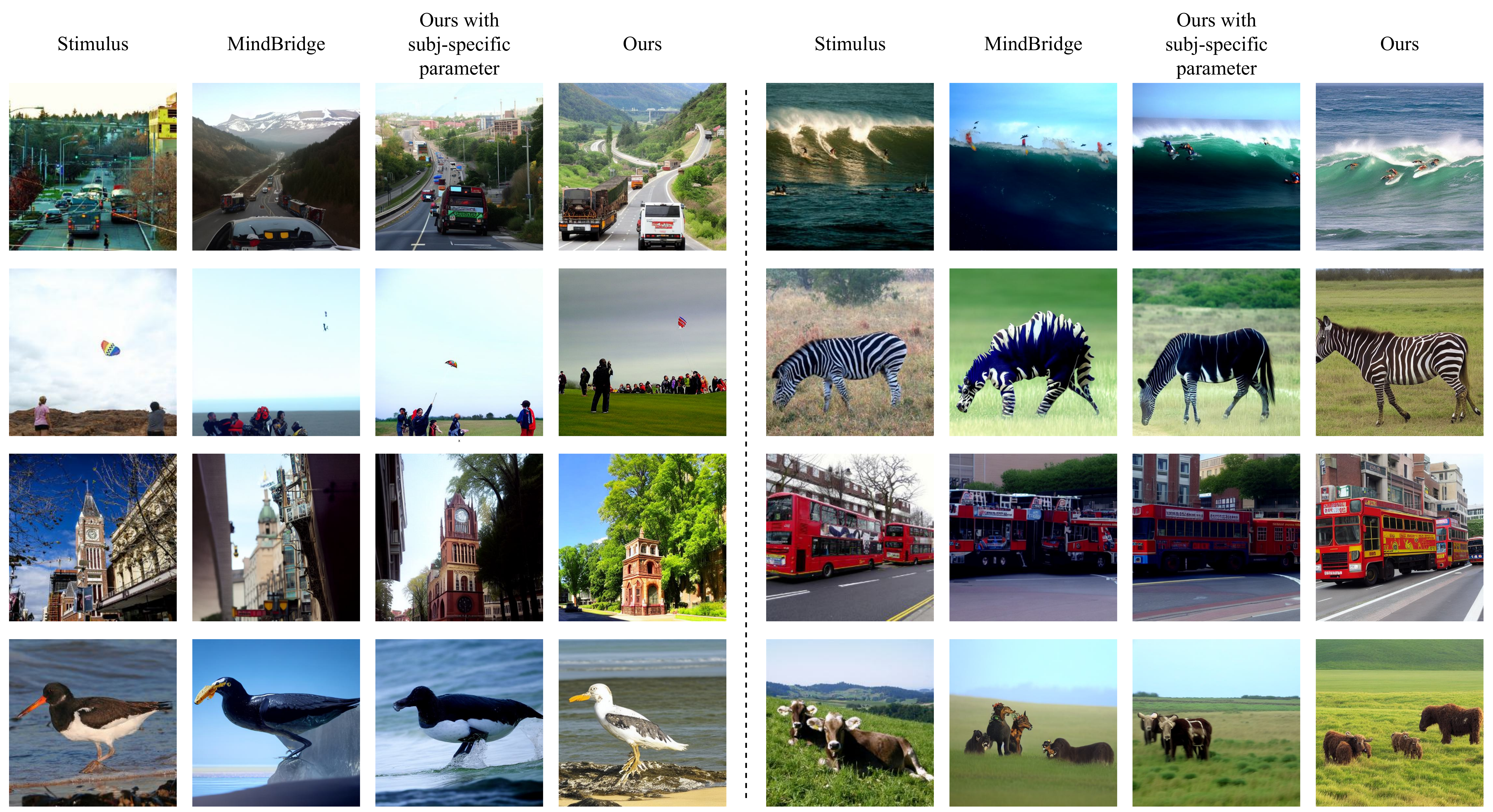}
\caption{Visualizations of different fMRI decoding methods, including our newly proposed UniBrain method, our UniBrain method using subject-specific parameters and MindBridge, on the NSD dataset. All of the fMRI signals are from subject 1.}
\label{fig_vis_fully_supervise}
\end{figure*}
 
\section{Experiments}
\label{sec:exp}

\begin{table*}[t]\centering
\scalebox{0.825}{
\begin{tabular}{lcccccccc}
\toprule
\multirow{2}{*}{Method} & \multicolumn{4}{c}{Low-Level} & \multicolumn{4}{c}{High-Level} \\ 
\cline{2-9} & PixCorr $\uparrow$ & SSIM $\uparrow$ & Alex(2) $\uparrow$ & Alex(5) $\uparrow$ & Incep $\uparrow$ & CLIP $\uparrow$ & EffNet-B $\downarrow$ & SwAV $\downarrow$ \\ 
\hline
Fully-supervise & .155 & .259 & 87.8\% & 95.5\% & 92.4\% & 94.0\% & .691 & .407 \\
MindBridge~\cite{wang2024mindbridge} & .031 & .132 & 50.7\% & 52.3\% & 50.8\% & 54.1\% & .976 & .674 \\
UniBrain (Ours) & \textbf{.038} & \textbf{.166} & \textbf{53.1\%} & \textbf{55.4\%} & \textbf{51.7\%} & \textbf{54.6\%} & \textbf{.975} & \textbf{.643} \\
\bottomrule
\end{tabular}
}
\caption{Quantitative results on our newly introduced generalization benchmark. We report the results of the fully supervised method, the reproduced MindBridge, and our UniBrain. All methods are trained using three out of four subjects and tested on the remaining one subject. Metrics are calculated as the average across four subjects. The best performance is highlighted in bold.}
\label{table_DG}
\end{table*}

\begin{figure}[t]
\centering
\includegraphics[width=0.9\linewidth]{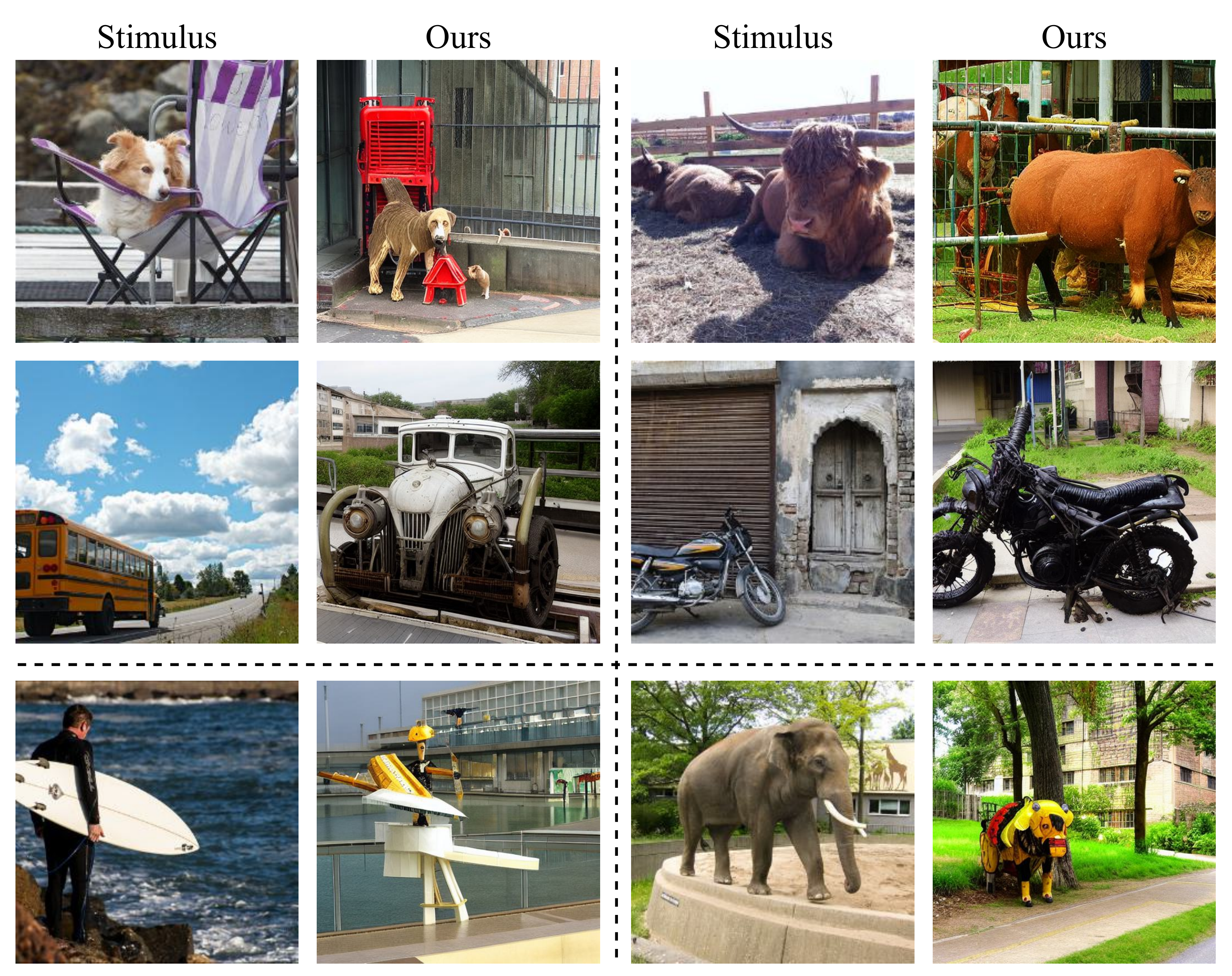}
\caption{Visualizations of the generalization of our newly proposed UniBrain on the NSD dataset, including both the successful cases (row 1 and row 2), and the failure cases (row 3). The model is trained using subjects 1, 2, 5, and tested on subject 7.}
\vspace{-4mm}
\label{fig_vis_DG}
\end{figure}

\begin{table*}[t]\centering
\scalebox{0.825}{
\begin{tabular}{cccccccccc}
\toprule
\multirow{2}{*}{MutAss} & \multirow{2}{*}{Adversarial} & \multicolumn{4}{c}{Low-Level} & \multicolumn{4}{c}{High-Level} \\ 
\cline{3-10} & & PixCorr $\uparrow$ & SSIM $\uparrow$ & Alex(2) $\uparrow$ & Alex(5) $\uparrow$ & Incep $\uparrow$ & CLIP $\uparrow$ & EffNet-B $\downarrow$ & SwAV $\downarrow$ \\ 
\hline
& & .149 & .223 & 84.2\% & 94.7\% & 91.2\% & 93.0\% & .745 & .446 \\
\ding{51} & & .148 & .253 & 88.1\% & 95.6\% & 92.2\% & 94.1\% & .686 & .407 \\
& \ding{51} & .139 & .239 & 85.9\% & 94.4\% & 90.3\% & 93.1\% & .745 & .432 \\
\ding{51} & \ding{51} & .155 & .259 & 87.8\% & 95.5\% & 92.4\% & 94.0\% & .691 & .407 \\
\bottomrule
\end{tabular}
}
\caption{Ablation study on the effectiveness of different components in our method, including the mutual assistance embedder (MutAss) and the adversarial subject-invariant feature extraction (Adversarial) module. All methods are trained and tested using four subjects. Metrics are calculated as the average across four subjects.}
\vspace{-3mm}
\label{ablation_module}
\end{table*}

\begin{table*}[t]\centering
\scalebox{0.825}{
\begin{tabular}{ccccccccccc}
\toprule
\multirow{2}{*}{Semantic} & \multirow{2}{*}{Geometric} & \multirow{2}{*}{Assist} & \multicolumn{4}{c}{Low-Level} & \multicolumn{4}{c}{High-Level} \\ 
\cline{4-11} & & & PixCorr $\uparrow$ & SSIM $\uparrow$ & Alex(2) $\uparrow$ & Alex(5) $\uparrow$ & Incep $\uparrow$ & CLIP $\uparrow$ & EffNet-B $\downarrow$ & SwAV $\downarrow$ \\ 
\hline
\ding{51} & & & .009 & .245 & 74.7\% & 87.5\% & 87.8\% & 88.5\% & .741 & .464 \\
& \ding{51} & & .149 & .223 & 84.2\% & 94.7\% & 91.2\% & 93.0\% & .745 & .446 \\
\ding{51} & \ding{51} & & .148 & .242 & 87.2\% & 95.3\% & 92.7\% &  93.9\% & .699 & .412 \\
\ding{51} & \ding{51} & \ding{51} & .148 & .253 & 88.1\% & 95.6\% & 92.2\% & 94.1\% & .686 & .407 \\
\bottomrule
\end{tabular}
}
\caption{Ablation study on the effectiveness of our mutual assistance embedder (MutAss), we compare the performance of different variants of MutAss, including using the semantic encoder (Semantic) only, using the geometric encoder (Geometric) only,  using both encoders and our mutual assistance (Assist) scheme. All methods are trained and tested using four subjects. Metrics are calculated as the average across four subjects. Note that the adversarial subject-invariant feature extraction (Adversarial) module is not adopted here.}
\label{ablation_MutAss}
\end{table*}

\begin{table*}[ht]\centering
\scalebox{0.825}{
\begin{tabular}{lccccccccc}
\toprule
\multirow{2}{*}{Loss} & \multicolumn{4}{c}{Low-Level} & \multicolumn{4}{c}{High-Level} \\ 
\cline{2-9} & PixCorr $\uparrow$ & SSIM $\uparrow$ & Alex(2) $\uparrow$ & Alex(5) $\uparrow$ & Incep $\uparrow$ & CLIP $\uparrow$ & EffNet-B $\downarrow$ & SwAV $\downarrow$ \\ 
\hline
UniBrain (w/o SoftCLIP) & .147 & .262 & 85.0\% & 94.0\% & 90.3\% & 92.2\% & .696 & .408 \\
UniBrain (w/o MSE) & .081 & .324 & 75.2\% & 81.7\% & 78.5\% & 78.4\% & .890 & .558 \\
UniBrain & .148 & .253 & 88.1\% & 95.6\% & 92.2\% & 94.1\% & .686 & .407 \\
\bottomrule
\end{tabular}
}
\caption{Ablation study on the effectiveness of different loss functions in our method, including the MSE loss and the SoftCLIP loss. All methods are trained and tested using four subjects. Metrics are calculated as the average across four subjects. Note that the adversarial subject-invariant feature extraction (Adversarial) module is not adopted here.}
\label{ablation_loss}
\end{table*}

\begin{table}[ht]\centering
\scalebox{0.525}{
\begin{tabular}{ccccccccccc}
\toprule
\multirow{2}{*}{$G$} & \multirow{2}{*}{$K$} & \multicolumn{4}{c}{Low-Level} & \multicolumn{4}{c}{High-Level} \\ 
\cline{3-10} & & PixCorr $\uparrow$ & SSIM $\uparrow$ & Alex(2) $\uparrow$ & Alex(5) $\uparrow$ & Incep $\uparrow$ & CLIP $\uparrow$ & EffNet-B $\downarrow$ & SwAV $\downarrow$ \\ 
\hline
256 & 16 & .129 & .239 & 83.7\% & 93.1\% & 88.3\% & 90.9\% & .754 & .451 \\
256 & 32 & .142 & .240 & 86.3\% & 94.6\% & 91.0\% & 92.9\% & .712 & .423 \\
256 & 64 & .139 & .240 & 86.8\% & 95.1\% & 91.5\% & 93.3\% &.704 & .420 \\
512 & 16 & .138 & .243 & 86.3\% & 94.4\% & 90.6\% & 92.8\% & .716 & .427 \\
512 & 32 & .155 & .259 & 87.8\% & 95.5\% & 92.4\% & 94.0\% & .691 & .407 \\
512 & 64 & .149 & .237 & 87.9\% & 95.8\% & 92.6\% & 94.3\% & .685 & .403 \\
1024 & 16 & .149 & .247 & 87.4\% & 95.3\% & 92.1\% & 94.0\% & .694 & .411 \\
1024 & 32 & .154 & .254 & 88.3\% & 96.1\% & 93.3\% & 94.7\% & .674 & .396 \\
\bottomrule
\end{tabular}
}
\caption{Ablation study on the group-based extractors in our method. We vary the number of the groups ($G$) and voxels $K$ within a group. All methods are trained and tested using four subjects. Metrics are calculated as the average across four subjects.}
\vspace{-3mm}
\label{ablation_group}
\end{table}

\subsection{Implementation Details}
\label{sec:exp_implement}
Following previous works~\cite{wang2024mindbridge, scotti2024reconstructing, takagi2023high, ozcelik2023natural}, we conduct our experiments on the widely used NSD dataset~\cite{allen2022massive} to verify the effectiveness of our method, and we adopt eight image quality evaluation metrics to quantify the performance of our method, \textit{i.e.}, we use PixCorr, SSIM~\cite{wang2004image}, AlexNet(2) and AlexNet(5)~\cite{krizhevsky2012imagenet} to evaluate low-level properties, and use Inception~\cite{szegedy2016rethinking}, CLIP~\cite{radford2021learning}, EffNet-B~\cite{tan2019efficientnet} and SwAV~\cite{caron2020unsupervised} to evaluate high-level properties. More details on implementation are listed in the supplementary materials.

\subsection{Experimental Results}
\label{sec:exp_performance}
We first compare the quantitative performance of our UniBrain with current SOTA methods in a fully supervised manner in Table~\ref{table_fully_supervise}. Considering current methods either train separate models for different subjects or a partially unified model utilizing subject-specific parameters, for a fair comparison, we also report the performance of our method using subject-specific parameters. In particular, we still use a unified embedded, but train separate extractors for different individuals, and we also adopt the low-level stream following the same setting as MindEye~\cite{scotti2024reconstructing}. It can be inferred from the table that our newly proposed UniBrain method can achieve better performance compared with previous partially unified methods like MindBridge. In addition, when introducing subject-specific parameters, our method surpasses MindBridge by a large margin in terms of all eight metrics, and can also achieve comparable performance with subject-specific methods like MindEye, indicating the great effectiveness of our method.
Besides, we also provide some visualization results of our method in Fig.~\ref{fig_vis_fully_supervise}. It can be inferred from the figure that our method with subject-specific parameters achieves excellent decoding performance in terms of both semantic consistency and image details, like the town road with trees, the bell tower with trees, the surfing people, etc. 
Additionally, the figure suggests that even without subject-specific parameters, our UniBrain model outperforms MindBridge, though it may perform slightly below UniBrain with subject-specific parameters under circumstances like counting the number of cows or identifying the colour of a bird.

In addition, considering that we target extracting cross-subject commonalities and encourage the model to be directly generalized to new subjects, we also report the performance of our method when training on three subjects and testing on the rest one subject in Table~\ref{table_DG} and the visualizations in Fig.~\ref{fig_vis_DG}. We reproduce MindBridge using a unified extractor to replace the subject-specific modules. It can be inferred from the table that our method outperforms MindBridge by a large margin. It can also be reasoned from the figure that the semantics of the reconstructed images using our method align well with the original stimuli, like the dog, the cow, and the motorcycle. However, failure cases apparently exist, like the man carrying the surfboard or the elephant, the semantics of the reconstructed images might be slightly different from the original stimuli, but we can still infer the similarities from a geometric view.

\textbf{Limitations and discussions:} The generalization performance of our method still lies far behind that of fully supervised methods. 
As is known, the generation of the brain is a complex process~\cite{stiles2010basics}, it is still unknown whether the cross-subject commonalities exist or whether cross-subject image reconstruction is possible. This topic still needs further research. 
Besides, the NSD dataset is collected using only a couple of subjects with only around ten thousand fMRI-to-image pairs, which is quite small compared with the training data of visual foundation models like SAM~\cite{kirillov2023segment, ravi2024sam} or DINO~\cite{caron2021emerging, oquab2023dinov2}. Whether a model trained using such little data is capable of capturing cross-subject and even population-level commonalities is not sure.
Therefore, we call on the community to collect more data to promote the research on general brain decoding, which should concern not only the stimuli diversity but also the individual diversity. 
In addition, whether the difficulty in capturing cross-subject commonalities due to the shortage of data or the limited power of the model is also unknown. Therefore, a strong model is also desired to better capture the cross-subject commonalities. 
Facing the issues, we come up with the generalization benchmark for brain decoding and we advocate for greater emphasis on cross-subject commonalities in generalized brain decoding. We hope such research may contribute to more practical utility in medicine and industry.

% \textbf{Limitations and discussions:} While promising, our method’s cross-subject OOD generalization performance remains below that of subject-specific approaches. Brain decoding is a complex process~\cite{stiles2010basics}, and the cross-subject commonalities need deeper investigation. Additionally, despite containing ten thousand fMRI-to-image pairs, the NSD dataset is limited in a few subjects, while the scale of the dataset is much smaller than the extensive datasets used to train visual foundation models like SAM~\cite{kirillov2023segment, ravi2024sam} and DINO~\cite{caron2021emerging, oquab2023dinov2}. It remains to be seen whether a model trained on such a dataset can adequately capture cross-subject or population-level patterns. We encourage the community to expand data collection efforts to further general brain decoding research, prioritizing both stimulus and individual diversity. With these considerations, we introduce a benchmark for generalization in brain decoding and advocate for a focus on cross-subject commonalities to support more practical applications in medicine and industry.

\subsection{Ablation Study}
\label{sec:exp_ablation}
% In this section, we single out the contributions of our designs, all methods are trained and tested using four subjects.

Recall that our UniBrain includes a group-based extractor, a mutual assistance embedder and a bilevel feature alignment scheme, where the group-based extractor is a compulsory component for cross-subject model designing, and the CLIP feature alignment is also compulsory for guiding reconstruction. Therefore, in Table~\ref{ablation_module}, we analyze the effectiveness of the adversarial subject-invariant feature extraction module and the mutual assistance embedder. It is clear from the table that our mutual assistance embedder contributes significantly to the brain decoding performance on both high-level and low-level metrics, indicating that our mutual assistance embedder better captures not only low-level geometric information but also high-level semantic information. Besides, the subject-invariant feature extraction scheme will also contribute to the cross-subject information extraction, leading to a better decoding performance.

Then we further delve into the intricate design of our mutual assistance embedder. Recall that our mutual assistance embedder includes a geometric embedder, a semantic embedder and a unified embedder. It can be inferred from Table~\ref{ablation_MutAss} that utilizing semantic embedder alone would lead to a very limited decoding performance, the main reason is that the semantic information will only provide an abstract concept of the stimuli, which will lose potential fine-grained details of the initial stimuli. On the contrary, the geometric embedder will extract fine-grained image-level geometric information, leading to a better decoding performance. However, the high-level information extraction capability of the geometric embedder is slightly insufficient. In consequence, combining the semantic embedder and the geometric embedder will bring a better decoding performance as both the semantic information and the geometric information can be utilized to guide the brain decoding. Finally, our mutual assistance scheme further enhances the high-level and low-level feature extraction ability as the semantic information can assist geometric information extraction by providing high-level semantic context and geometric information can also enhance semantic extraction by contributing fine-grained image-level details.

In addition, we provide the ablation on the loss functions we use in our UniBrain method in Table~\ref{ablation_loss}, which includes an MSE loss and a SoftCLIP loss. It can be inferred from the table that solely using MSE loss for feature alignment can bring a satisfying decoding performance. Combining the SoftCLIP loss with the MSE loss will not only enable the feature to be aligned well with the corresponding CLIP features but also enable the model to learn the relationship between subjects, leading to better decoding performance.

Finally, we also analyze the robustness of our group-based extractor in Table~\ref{ablation_group}. It can be inferred that varying the number of groups $G$ and the number of voxels within a group $K$ with a fixed $G \times K$ will not influence the decoding performance severely, indicating the robustness of our method. However, a small $G \times K$ will lead to a drop in the decoding performance, mainly due to information loss during the grouping stage. It can be reasoned from the table that the fewer voxels we utilize, the worse the decoding performance. Besides, a large $G \times K$ will lead to better decoding performance, but will also introduce memory usage. Considering the performance and the cost, we select $G=512$ and $K=32$ in our experiments. 
% More ablation studies will be provided in the supplementary materials.

\section{Conclusion}
\label{sec:conclusion}

In this work, we have presented a new unified model for cross-subject brain decoding. In particular, we propose a group-based extractor, a mutual assistance embedder and a bilevel feature alignment scheme to encourage the model to capture cross-subject commonalities and can be directly adopted on new subjects. 
Extensive experiments on the NSD benchmark have validated the effectiveness of our approach, where our unified method achieves comparable performance with state-of-the-art subject-specific methods.
We have also developed a new generalization benchmark for brain decoding.
We encourage the community to focus on increasing the diversity of the data and improving the generalization of the model to emphasise cross-subject commonalities in generalized brain decoding.

{
    \small
    \bibliographystyle{ieeenat_fullname}
    \bibliography{main}

\begin{thebibliography}{58}
\providecommand{\natexlab}[1]{#1}
\providecommand{\url}[1]{\texttt{#1}}
\expandafter\ifx\csname urlstyle\endcsname\relax
  \providecommand{\doi}[1]{doi: #1}\else
  \providecommand{\doi}{doi: \begingroup \urlstyle{rm}\Url}\fi

\bibitem[Alayrac et~al.(2022)Alayrac, Donahue, Luc, Miech, Barr, Hasson, Lenc, Mensch, Millican, Reynolds, et~al.]{alayrac2022flamingo}
Jean-Baptiste Alayrac, Jeff Donahue, Pauline Luc, Antoine Miech, Iain Barr, Yana Hasson, Karel Lenc, Arthur Mensch, Katherine Millican, Malcolm Reynolds, et~al.
\newblock Flamingo: a visual language model for few-shot learning.
\newblock \emph{Advances in neural information processing systems}, 35:\penalty0 23716--23736, 2022.

\bibitem[Allen et~al.(2022)Allen, St-Yves, Wu, Breedlove, Prince, Dowdle, Nau, Caron, Pestilli, Charest, et~al.]{allen2022massive}
Emily~J Allen, Ghislain St-Yves, Yihan Wu, Jesse~L Breedlove, Jacob~S Prince, Logan~T Dowdle, Matthias Nau, Brad Caron, Franco Pestilli, Ian Charest, et~al.
\newblock A massive 7t fmri dataset to bridge cognitive neuroscience and artificial intelligence.
\newblock \emph{Nature neuroscience}, 25\penalty0 (1):\penalty0 116--126, 2022.

\bibitem[Amaro~Jr and Barker(2006)]{amaro2006study}
Edson Amaro~Jr and Gareth~J Barker.
\newblock Study design in fmri: basic principles.
\newblock \emph{Brain and cognition}, 60\penalty0 (3):\penalty0 220--232, 2006.

\bibitem[Ba(2016)]{ba2016layer}
Jimmy~Lei Ba.
\newblock Layer normalization.
\newblock \emph{arXiv preprint arXiv:1607.06450}, 2016.

\bibitem[Beliy et~al.(2019)Beliy, Gaziv, Hoogi, Strappini, Golan, and Irani]{beliy2019voxels}
Roman Beliy, Guy Gaziv, Assaf Hoogi, Francesca Strappini, Tal Golan, and Michal Irani.
\newblock From voxels to pixels and back: Self-supervision in natural-image reconstruction from fmri.
\newblock \emph{Advances in Neural Information Processing Systems}, 32, 2019.

\bibitem[Caron et~al.(2020)Caron, Misra, Mairal, Goyal, Bojanowski, and Joulin]{caron2020unsupervised}
Mathilde Caron, Ishan Misra, Julien Mairal, Priya Goyal, Piotr Bojanowski, and Armand Joulin.
\newblock Unsupervised learning of visual features by contrasting cluster assignments.
\newblock \emph{Advances in neural information processing systems}, 33:\penalty0 9912--9924, 2020.

\bibitem[Caron et~al.(2021)Caron, Touvron, Misra, J{\'e}gou, Mairal, Bojanowski, and Joulin]{caron2021emerging}
Mathilde Caron, Hugo Touvron, Ishan Misra, Herv{\'e} J{\'e}gou, Julien Mairal, Piotr Bojanowski, and Armand Joulin.
\newblock Emerging properties in self-supervised vision transformers.
\newblock In \emph{Proceedings of the IEEE/CVF international conference on computer vision}, pages 9650--9660, 2021.

\bibitem[{\c{C}}elik et~al.(2019){\c{C}}elik, Dar, Y{\i}lmaz, Kele{\c{s}}, and {\c{C}}ukur]{ccelik2019spatially}
Emin {\c{C}}elik, Salman Ul~Hassan Dar, {\"O}zg{\"u}r Y{\i}lmaz, {\"U}mit Kele{\c{s}}, and Tolga {\c{C}}ukur.
\newblock Spatially informed voxelwise modeling for naturalistic fmri experiments.
\newblock \emph{NeuroImage}, 186:\penalty0 741--757, 2019.

\bibitem[Chen et~al.(2024)Chen, Qi, Wang, and Pan]{chen2024mind}
Jiaxuan Chen, Yu Qi, Yueming Wang, and Gang Pan.
\newblock Mind artist: Creating artistic snapshots with human thought.
\newblock In \emph{Proceedings of the IEEE/CVF Conference on Computer Vision and Pattern Recognition}, pages 27207--27217, 2024.

\bibitem[Engel et~al.(1994)Engel, Rumelhart, Wandell, Lee, Glover, Chichilnisky, Shadlen, et~al.]{engel1994fmri}
Stephen~A Engel, David~E Rumelhart, Brian~A Wandell, Adrian~T Lee, Gary~H Glover, Eduardo-Jose Chichilnisky, Michael~N Shadlen, et~al.
\newblock fmri of human visual cortex.
\newblock \emph{Nature}, 369\penalty0 (6481):\penalty0 525--525, 1994.

\bibitem[Fang et~al.(2024)Fang, Zheng, and Pan]{fang2024alleviating}
Tao Fang, Qian Zheng, and Gang Pan.
\newblock Alleviating the semantic gap for generalized fmri-to-image reconstruction.
\newblock \emph{Advances in Neural Information Processing Systems}, 36, 2024.

\bibitem[Finn et~al.(2017)Finn, Scheinost, Finn, Shen, Papademetris, and Constable]{finn2017can}
Emily~S Finn, Dustin Scheinost, Daniel~M Finn, Xilin Shen, Xenophon Papademetris, and R~Todd Constable.
\newblock Can brain state be manipulated to emphasize individual differences in functional connectivity?
\newblock \emph{NeuroImage}, 160:\penalty0 140--151, 2017.

\bibitem[Ganin et~al.(2016)Ganin, Ustinova, Ajakan, Germain, Larochelle, Laviolette, March, and Lempitsky]{ganin2016domain}
Yaroslav Ganin, Evgeniya Ustinova, Hana Ajakan, Pascal Germain, Hugo Larochelle, Fran{\c{c}}ois Laviolette, Mario March, and Victor Lempitsky.
\newblock Domain-adversarial training of neural networks.
\newblock \emph{Journal of machine learning research}, 17\penalty0 (59):\penalty0 1--35, 2016.

\bibitem[Geerligs et~al.(2015)Geerligs, Rubinov, Henson, et~al.]{geerligs2015state}
Linda Geerligs, Mikail Rubinov, Richard~N Henson, et~al.
\newblock State and trait components of functional connectivity: individual differences vary with mental state.
\newblock \emph{Journal of Neuroscience}, 35\penalty0 (41):\penalty0 13949--13961, 2015.

\bibitem[Haynes and Rees(2005)]{haynes2005predicting}
John-Dylan Haynes and Geraint Rees.
\newblock Predicting the orientation of invisible stimuli from activity in human primary visual cortex.
\newblock \emph{Nature neuroscience}, 8\penalty0 (5):\penalty0 686--691, 2005.

\bibitem[Haynes and Rees(2006)]{haynes2006decoding}
John-Dylan Haynes and Geraint Rees.
\newblock Decoding mental states from brain activity in humans.
\newblock \emph{Nature reviews neuroscience}, 7\penalty0 (7):\penalty0 523--534, 2006.

\bibitem[Hendrycks and Gimpel(2016)]{hendrycks2016gaussian}
Dan Hendrycks and Kevin Gimpel.
\newblock Gaussian error linear units (gelus).
\newblock \emph{arXiv preprint arXiv:1606.08415}, 2016.

\bibitem[Ho et~al.(2020)Ho, Jain, and Abbeel]{ho2020denoising}
Jonathan Ho, Ajay Jain, and Pieter Abbeel.
\newblock Denoising diffusion probabilistic models.
\newblock \emph{Advances in neural information processing systems}, 33:\penalty0 6840--6851, 2020.

\bibitem[Horikawa and Kamitani(2017)]{horikawa2017generic}
Tomoyasu Horikawa and Yukiyasu Kamitani.
\newblock Generic decoding of seen and imagined objects using hierarchical visual features.
\newblock \emph{Nature communications}, 8\penalty0 (1):\penalty0 15037, 2017.

\bibitem[Huo et~al.(2024)Huo, Wang, Qian, Wang, Li, Feng, and Fu]{huo2024neuropictor}
Jingyang Huo, Yikai Wang, Xuelin Qian, Yun Wang, Chong Li, Jianfeng Feng, and Yanwei Fu.
\newblock Neuropictor: Refining fmri-to-image reconstruction via multi-individual pretraining and multi-level modulation.
\newblock \emph{arXiv preprint arXiv:2403.18211}, 2024.

\bibitem[Kamitani and Tong(2005)]{kamitani2005decoding}
Yukiyasu Kamitani and Frank Tong.
\newblock Decoding the visual and subjective contents of the human brain.
\newblock \emph{Nature neuroscience}, 8\penalty0 (5):\penalty0 679--685, 2005.

\bibitem[Kay and Gallant(2009)]{kay2009can}
Kendrick~N Kay and Jack~L Gallant.
\newblock I can see what you see.
\newblock \emph{Nature neuroscience}, 12\penalty0 (3):\penalty0 245--245, 2009.

\bibitem[Kay et~al.(2008)Kay, Naselaris, Prenger, and Gallant]{kay2008identifying}
Kendrick~N Kay, Thomas Naselaris, Ryan~J Prenger, and Jack~L Gallant.
\newblock Identifying natural images from human brain activity.
\newblock \emph{Nature}, 452\penalty0 (7185):\penalty0 352--355, 2008.

\bibitem[Kirillov et~al.(2023)Kirillov, Mintun, Ravi, Mao, Rolland, Gustafson, Xiao, Whitehead, Berg, Lo, et~al.]{kirillov2023segment}
Alexander Kirillov, Eric Mintun, Nikhila Ravi, Hanzi Mao, Chloe Rolland, Laura Gustafson, Tete Xiao, Spencer Whitehead, Alexander~C Berg, Wan-Yen Lo, et~al.
\newblock Segment anything.
\newblock In \emph{Proceedings of the IEEE/CVF International Conference on Computer Vision}, pages 4015--4026, 2023.

\bibitem[Krizhevsky et~al.(2012)Krizhevsky, Sutskever, and Hinton]{krizhevsky2012imagenet}
Alex Krizhevsky, Ilya Sutskever, and Geoffrey~E Hinton.
\newblock Imagenet classification with deep convolutional neural networks.
\newblock \emph{Advances in neural information processing systems}, 25, 2012.

\bibitem[Lin et~al.(2022)Lin, Sprague, and Singh]{lin2022mind}
Sikun Lin, Thomas Sprague, and Ambuj~K Singh.
\newblock Mind reader: Reconstructing complex images from brain activities.
\newblock \emph{Advances in Neural Information Processing Systems}, 35:\penalty0 29624--29636, 2022.

\bibitem[Lin et~al.(2014)Lin, Maire, Belongie, Hays, Perona, Ramanan, Doll{\'a}r, and Zitnick]{lin2014microsoft}
Tsung-Yi Lin, Michael Maire, Serge Belongie, James Hays, Pietro Perona, Deva Ramanan, Piotr Doll{\'a}r, and C~Lawrence Zitnick.
\newblock Microsoft coco: Common objects in context.
\newblock In \emph{Computer Vision--ECCV 2014: 13th European Conference, Zurich, Switzerland, September 6-12, 2014, Proceedings, Part V 13}, pages 740--755. Springer, 2014.

\bibitem[Loshchilov(2017)]{loshchilov2017decoupled}
I Loshchilov.
\newblock Decoupled weight decay regularization.
\newblock \emph{arXiv preprint arXiv:1711.05101}, 2017.

\bibitem[Lu et~al.(2023)Lu, Du, Zhou, Wang, and He]{lu2023minddiffuser}
Yizhuo Lu, Changde Du, Qiongyi Zhou, Dianpeng Wang, and Huiguang He.
\newblock Minddiffuser: Controlled image reconstruction from human brain activity with semantic and structural diffusion.
\newblock In \emph{Proceedings of the 31st ACM International Conference on Multimedia}, pages 5899--5908, 2023.

\bibitem[Luo et~al.(2024)Luo, Henderson, Wehbe, and Tarr]{luo2024brain}
Andrew Luo, Maggie Henderson, Leila Wehbe, and Michael Tarr.
\newblock Brain diffusion for visual exploration: Cortical discovery using large scale generative models.
\newblock \emph{Advances in Neural Information Processing Systems}, 36, 2024.

\bibitem[Miyawaki et~al.(2008)Miyawaki, Uchida, Yamashita, Sato, Morito, Tanabe, Sadato, and Kamitani]{miyawaki2008visual}
Yoichi Miyawaki, Hajime Uchida, Okito Yamashita, Masa-aki Sato, Yusuke Morito, Hiroki~C Tanabe, Norihiro Sadato, and Yukiyasu Kamitani.
\newblock Visual image reconstruction from human brain activity using a combination of multiscale local image decoders.
\newblock \emph{Neuron}, 60\penalty0 (5):\penalty0 915--929, 2008.

\bibitem[Mueller et~al.(2013)Mueller, Wang, Fox, Yeo, Sepulcre, Sabuncu, Shafee, Lu, and Liu]{mueller2013individual}
Sophia Mueller, Danhong Wang, Michael~D Fox, BT~Thomas Yeo, Jorge Sepulcre, Mert~R Sabuncu, Rebecca Shafee, Jie Lu, and Hesheng Liu.
\newblock Individual variability in functional connectivity architecture of the human brain.
\newblock \emph{Neuron}, 77\penalty0 (3):\penalty0 586--595, 2013.

\bibitem[Naselaris et~al.(2011)Naselaris, Kay, Nishimoto, and Gallant]{naselaris2011encoding}
Thomas Naselaris, Kendrick~N Kay, Shinji Nishimoto, and Jack~L Gallant.
\newblock Encoding and decoding in fmri.
\newblock \emph{Neuroimage}, 56\penalty0 (2):\penalty0 400--410, 2011.

\bibitem[Nishimoto et~al.(2011)Nishimoto, Vu, Naselaris, Benjamini, Yu, and Gallant]{nishimoto2011reconstructing}
Shinji Nishimoto, An~T Vu, Thomas Naselaris, Yuval Benjamini, Bin Yu, and Jack~L Gallant.
\newblock Reconstructing visual experiences from brain activity evoked by natural movies.
\newblock \emph{Current biology}, 21\penalty0 (19):\penalty0 1641--1646, 2011.

\bibitem[Oquab et~al.(2023)Oquab, Darcet, Moutakanni, Vo, Szafraniec, Khalidov, Fernandez, Haziza, Massa, El-Nouby, et~al.]{oquab2023dinov2}
Maxime Oquab, Timoth{\'e}e Darcet, Th{\'e}o Moutakanni, Huy Vo, Marc Szafraniec, Vasil Khalidov, Pierre Fernandez, Daniel Haziza, Francisco Massa, Alaaeldin El-Nouby, et~al.
\newblock Dinov2: Learning robust visual features without supervision.
\newblock \emph{arXiv preprint arXiv:2304.07193}, 2023.

\bibitem[Ozcelik and VanRullen(2023)]{ozcelik2023natural}
Furkan Ozcelik and Rufin VanRullen.
\newblock Natural scene reconstruction from fmri signals using generative latent diffusion.
\newblock \emph{Scientific Reports}, 13\penalty0 (1):\penalty0 15666, 2023.

\bibitem[Peebles and Xie(2023)]{peebles2023scalable}
William Peebles and Saining Xie.
\newblock Scalable diffusion models with transformers.
\newblock In \emph{Proceedings of the IEEE/CVF International Conference on Computer Vision}, pages 4195--4205, 2023.

\bibitem[Radford et~al.(2021)Radford, Kim, Hallacy, Ramesh, Goh, Agarwal, Sastry, Askell, Mishkin, Clark, et~al.]{radford2021learning}
Alec Radford, Jong~Wook Kim, Chris Hallacy, Aditya Ramesh, Gabriel Goh, Sandhini Agarwal, Girish Sastry, Amanda Askell, Pamela Mishkin, Jack Clark, et~al.
\newblock Learning transferable visual models from natural language supervision.
\newblock In \emph{International conference on machine learning}, pages 8748--8763. PMLR, 2021.

\bibitem[Ravi et~al.(2024)Ravi, Gabeur, Hu, Hu, Ryali, Ma, Khedr, R{\"a}dle, Rolland, Gustafson, et~al.]{ravi2024sam}
Nikhila Ravi, Valentin Gabeur, Yuan-Ting Hu, Ronghang Hu, Chaitanya Ryali, Tengyu Ma, Haitham Khedr, Roman R{\"a}dle, Chloe Rolland, Laura Gustafson, et~al.
\newblock Sam 2: Segment anything in images and videos.
\newblock \emph{arXiv preprint arXiv:2408.00714}, 2024.

\bibitem[Rombach et~al.(2022)Rombach, Blattmann, Lorenz, Esser, and Ommer]{rombach2022high}
Robin Rombach, Andreas Blattmann, Dominik Lorenz, Patrick Esser, and Bj{\"o}rn Ommer.
\newblock High-resolution image synthesis with latent diffusion models.
\newblock In \emph{Proceedings of the IEEE/CVF conference on computer vision and pattern recognition}, pages 10684--10695, 2022.

\bibitem[Ruffle et~al.(2024)Ruffle, Gray, Mohinta, Pombo, Kaul, Hyare, Rees, and Nachev]{ruffle2024computational}
James~K Ruffle, Robert~J Gray, Samia Mohinta, Guilherme Pombo, Chaitanya Kaul, Harpreet Hyare, Geraint Rees, and Parashkev Nachev.
\newblock Computational limits to the legibility of the imaged human brain.
\newblock \emph{NeuroImage}, 291:\penalty0 120600, 2024.

\bibitem[Scotti et~al.(2024{\natexlab{a}})Scotti, Banerjee, Goode, Shabalin, Nguyen, Dempster, Verlinde, Yundler, Weisberg, Norman, et~al.]{scotti2024reconstructing}
Paul Scotti, Atmadeep Banerjee, Jimmie Goode, Stepan Shabalin, Alex Nguyen, Aidan Dempster, Nathalie Verlinde, Elad Yundler, David Weisberg, Kenneth Norman, et~al.
\newblock Reconstructing the mind's eye: fmri-to-image with contrastive learning and diffusion priors.
\newblock \emph{Advances in Neural Information Processing Systems}, 36, 2024{\natexlab{a}}.

\bibitem[Scotti et~al.(2024{\natexlab{b}})Scotti, Tripathy, Villanueva, Kneeland, Chen, Narang, Santhirasegaran, Xu, Naselaris, Norman, et~al.]{scotti2024mindeye2}
Paul~S Scotti, Mihir Tripathy, Cesar Kadir~Torrico Villanueva, Reese Kneeland, Tong Chen, Ashutosh Narang, Charan Santhirasegaran, Jonathan Xu, Thomas Naselaris, Kenneth~A Norman, et~al.
\newblock Mindeye2: Shared-subject models enable fmri-to-image with 1 hour of data.
\newblock \emph{arXiv preprint arXiv:2403.11207}, 2024{\natexlab{b}}.

\bibitem[Shen et~al.(2019{\natexlab{a}})Shen, Dwivedi, Majima, Horikawa, and Kamitani]{shen2019end}
Guohua Shen, Kshitij Dwivedi, Kei Majima, Tomoyasu Horikawa, and Yukiyasu Kamitani.
\newblock End-to-end deep image reconstruction from human brain activity.
\newblock \emph{Frontiers in computational neuroscience}, 13:\penalty0 432276, 2019{\natexlab{a}}.

\bibitem[Shen et~al.(2019{\natexlab{b}})Shen, Horikawa, Majima, and Kamitani]{shen2019deep}
Guohua Shen, Tomoyasu Horikawa, Kei Majima, and Yukiyasu Kamitani.
\newblock Deep image reconstruction from human brain activity.
\newblock \emph{PLoS computational biology}, 15\penalty0 (1):\penalty0 e1006633, 2019{\natexlab{b}}.

\bibitem[Shirakawa et~al.(2024)Shirakawa, Nagano, Tanaka, Aoki, Majima, Muraki, and Kamitani]{shirakawa2024spurious}
Ken Shirakawa, Yoshihiro Nagano, Misato Tanaka, Shuntaro~C Aoki, Kei Majima, Yusuke Muraki, and Yukiyasu Kamitani.
\newblock Spurious reconstruction from brain activity.
\newblock \emph{arXiv preprint arXiv:2405.10078}, 2024.

\bibitem[Srivastava et~al.(2014)Srivastava, Hinton, Krizhevsky, Sutskever, and Salakhutdinov]{srivastava2014dropout}
Nitish Srivastava, Geoffrey Hinton, Alex Krizhevsky, Ilya Sutskever, and Ruslan Salakhutdinov.
\newblock Dropout: a simple way to prevent neural networks from overfitting.
\newblock \emph{The journal of machine learning research}, 15\penalty0 (1):\penalty0 1929--1958, 2014.

\bibitem[Stiles and Jernigan(2010)]{stiles2010basics}
Joan Stiles and Terry~L Jernigan.
\newblock The basics of brain development.
\newblock \emph{Neuropsychology review}, 20\penalty0 (4):\penalty0 327--348, 2010.

\bibitem[Szegedy et~al.(2016)Szegedy, Vanhoucke, Ioffe, Shlens, and Wojna]{szegedy2016rethinking}
Christian Szegedy, Vincent Vanhoucke, Sergey Ioffe, Jon Shlens, and Zbigniew Wojna.
\newblock Rethinking the inception architecture for computer vision.
\newblock In \emph{Proceedings of the IEEE conference on computer vision and pattern recognition}, pages 2818--2826, 2016.

\bibitem[Takagi and Nishimoto(2023)]{takagi2023high}
Yu Takagi and Shinji Nishimoto.
\newblock High-resolution image reconstruction with latent diffusion models from human brain activity.
\newblock In \emph{Proceedings of the IEEE/CVF Conference on Computer Vision and Pattern Recognition}, pages 14453--14463, 2023.

\bibitem[Tan and Le(2019)]{tan2019efficientnet}
Mingxing Tan and Quoc Le.
\newblock Efficientnet: Rethinking model scaling for convolutional neural networks.
\newblock In \emph{International conference on machine learning}, pages 6105--6114. PMLR, 2019.

\bibitem[Vaswani(2017)]{vaswani2017attention}
A Vaswani.
\newblock Attention is all you need.
\newblock \emph{Advances in Neural Information Processing Systems}, 2017.

\bibitem[Vinje and Gallant(2000)]{vinje2000sparse}
William~E Vinje and Jack~L Gallant.
\newblock Sparse coding and decorrelation in primary visual cortex during natural vision.
\newblock \emph{Science}, 287\penalty0 (5456):\penalty0 1273--1276, 2000.

\bibitem[Wang et~al.(2024)Wang, Liu, Tan, and Wang]{wang2024mindbridge}
Shizun Wang, Songhua Liu, Zhenxiong Tan, and Xinchao Wang.
\newblock Mindbridge: A cross-subject brain decoding framework.
\newblock In \emph{Proceedings of the IEEE/CVF Conference on Computer Vision and Pattern Recognition}, pages 11333--11342, 2024.

\bibitem[Wang et~al.(2004)Wang, Bovik, Sheikh, and Simoncelli]{wang2004image}
Zhou Wang, Alan~C Bovik, Hamid~R Sheikh, and Eero~P Simoncelli.
\newblock Image quality assessment: from error visibility to structural similarity.
\newblock \emph{IEEE transactions on image processing}, 13\penalty0 (4):\penalty0 600--612, 2004.

\bibitem[Xia et~al.(2025)Xia, de~Charette, Oztireli, and Xue]{xia2025umbrae}
Weihao Xia, Raoul de Charette, Cengiz Oztireli, and Jing-Hao Xue.
\newblock Umbrae: Unified multimodal brain decoding.
\newblock In \emph{European Conference on Computer Vision}, pages 242--259. Springer, 2025.

\bibitem[Xu et~al.(2023)Xu, Wang, Zhang, Wang, and Shi]{xu2023versatile}
Xingqian Xu, Zhangyang Wang, Gong Zhang, Kai Wang, and Humphrey Shi.
\newblock Versatile diffusion: Text, images and variations all in one diffusion model.
\newblock In \emph{Proceedings of the IEEE/CVF International Conference on Computer Vision}, pages 7754--7765, 2023.

\bibitem[Zeng et~al.(2024)Zeng, Li, Liu, Gao, Jiang, Tang, Hu, Liu, and Zhang]{zeng2024controllable}
Bohan Zeng, Shanglin Li, Xuhui Liu, Sicheng Gao, Xiaolong Jiang, Xu Tang, Yao Hu, Jianzhuang Liu, and Baochang Zhang.
\newblock Controllable mind visual diffusion model.
\newblock In \emph{Proceedings of the AAAI Conference on Artificial Intelligence}, pages 6935--6943, 2024.

\end{thebibliography}
}

\clearpage
\setcounter{page}{1}
\maketitlesupplementary

\subsection{Implementation Details}
\label{sec:supp_implement}
In this section, we first describe the dataset we use, and then we give more details about our implementation. 

The Natural Scenes Dataset (NSD) dataset~\cite{allen2022massive} is a publicly available brain decoding dataset, the fMRI signals are collected from eight subjects using images from MSCOCO~\cite{lin2014microsoft} as stimuli using a 7-Tesla fMRI scanner. Following previous works~\cite{takagi2023high, scotti2024reconstructing, wang2024mindbridge}, we adopt data from four of the eight subjects who completed all imaging sessions, \textit{i.e.}, subj01, subj02, subj05, and subj07, as well as the the same standardized train/test splits. Note that we adopt the pre-processed fMRI data from MindEye~\cite{scotti2024reconstructing}.

\begin{table}[b]\centering
\scalebox{0.525}{
\begin{tabular}{lccccccccc}
\toprule
\multirow{2}{*}{Depth} & \multicolumn{4}{c}{Low-Level} & \multicolumn{4}{c}{High-Level} \\ 
\cline{2-9} & PixCorr $\uparrow$ & SSIM $\uparrow$ & Alex(2) $\uparrow$ & Alex(5) $\uparrow$ & Incep $\uparrow$ & CLIP $\uparrow$ & EffNet-B $\downarrow$ & SwAV $\downarrow$ \\ 
\hline
1 & .139 & .242 & 86.9\% & 94.9\% & 90.7\% & 93.0\% & .730 & .438 \\
2 & .155 & .259 & 87.8\% & 95.5\% & 92.4\% & 94.0\% & .691 & .407 \\
4 & .156 & .260 & 88.1\% & 95.6\% & 92.2\% & 93.9\% & .681 & .401 \\
6 & .151 & .257 & 88.1\% & 95.3\% & 92.0\% & 94.1\% & .681 & .401 \\
\bottomrule
\end{tabular}
}
\caption{Ablation study on the mutual assistance embedders in our method. We vary the number of the depth of the embedders. All methods are trained and tested using four subjects. Metrics are calculated as the average across four subjects.}
\label{ablation_depth}
\end{table}

We use one linear layer with an output dimension of 32 as our group-based local extractor, and another linear layer with an output dimension of 768 as our group-based global extractor. The projector consists of a linear layer followed by a LayerNorm layer~\cite{ba2016layer} and a drop-out operation~\cite{srivastava2014dropout} with a ratio of 0.5. The embedder is a Transformer architecture following~\cite{alayrac2022flamingo}. The subject discriminator consists of two linear layers, a LayerNorm layer, a GELU function~\cite{hendrycks2016gaussian} and a drop-out operation with a ratio of 0.5. 

The weights of the loss functions $\lambda_0$, $\lambda_1$ and $\lambda_2$ are set as 1.0, 1e5, 1.0, respectively. Note that we set $\lambda_1$ as 1e5 mainly to make sure that the value of different losses remains in a similar order of magnitude. 
We set the temperature parameter of the SoftCLIP loss $\tau$ as 0.005, and we follow DANN~\cite{ganin2016domain} to set the hyperparameter $\alpha$. We set the number of groups $G$ as 512 and the voxels within a group $K$ as 32. We set the dimension of the local representations $D_l$ as 32, and the dimension of the global brain representation $B_b$ as 768. We set the number of mapped tokens $T$ as 512. Besides, the number of CLIP image embeddings $T_g$ is 257 and the number of CLIP text embeddings $T_s$ is 77. 
Following previous work~\cite{wang2024mindbridge}, we set the text-to-image ratio of the Versatile Diffusion as 0.5, and the guidance scale is set as 5.0. The number of steps of the diffusion process is set as 20. In addition, the ViT-L/14 variant of CLIP is adopted here as the CLIP encoder.

We use an AdamW optimizer~\cite{loshchilov2017decoupled} and a cycle scheduler, with a maximum learning rate of 1.0e-4. We train the model for 600 epochs using four NVIDIA A100 GPUs with a memory size of 40G, and the batch size is set as 24 on each GPU.

\section{More Ablation Studies}
\label{sec:supp_ablation}
In this section, we conduct more ablation studies to verify the effectiveness of our method. 

\begin{table}[t]\centering
\scalebox{0.525}{
\begin{tabular}{lccccccccc}
\toprule
\multirow{2}{*}{Type} & \multicolumn{4}{c}{Low-Level} & \multicolumn{4}{c}{High-Level} \\ 
\cline{2-9} & PixCorr $\uparrow$ & SSIM $\uparrow$ & Alex(2) $\uparrow$ & Alex(5) $\uparrow$ & Incep $\uparrow$ & CLIP $\uparrow$ & EffNet-B $\downarrow$ & SwAV $\downarrow$ \\ 
\hline
None & .148 & .253 & 88.1\% & 95.6\% & 92.2\% & 94.1\% & .686 & .407 \\
Linear & .147 & .249 & 87.3\% & 95.6\% & 92.4\% & 93.6\% & .694 & .412 \\
Non-linear (2L) & .155 & .259 & 87.8\% & 95.5\% & 92.4\% & 94.0\% & .691 & .407 \\
Non-linear (3L) & .147 & .241 & 87.7\% & 95.8\% & 92.7\% & 94.2\% & .685 & .405 \\
\bottomrule
\end{tabular}
}
\caption{Ablation study on the subject discriminator in our method,. We vary the design of the discriminator, including a linear discriminator, a two-layer non-linear discriminator and a three-layer non-linear discriminator. All methods are trained and tested using four subjects. Metrics are calculated as the average across four subjects.}
\label{ablation_discriminator}
\end{table}

\begin{table}[t]\centering
\scalebox{0.525}{
\begin{tabular}{lccccccccc}
\toprule
\multirow{2}{*}{$\lambda_1$} & \multicolumn{4}{c}{Low-Level} & \multicolumn{4}{c}{High-Level} \\ 
\cline{2-9} & PixCorr $\uparrow$ & SSIM $\uparrow$ & Alex(2) $\uparrow$ & Alex(5) $\uparrow$ & Incep $\uparrow$ & CLIP $\uparrow$ & EffNet-B $\downarrow$ & SwAV $\downarrow$ \\ 
\hline
1e4 & .107 & .235 & 82.9\% & 92.9\% & 90.1\% & 91.9\% & .773 & .474 \\
2.5e4 & .130 & .236 & 86.1\% & 94.8\% & 92.1\% & 93.7\% & .728 & .444 \\
5e4 & .142 & .238 & 87.1\% & 95.5\% & 92.6\% & 94.1\% & .706 & .423 \\
1e5 & .155 & .259 & 87.8\% & 95.5\% & 92.4\% & 94.0\% & .691 & .407 \\
2e5 & .153 & .259 & 87.6\% & 95.3\% & 91.8\% & 93.5\% & .687 & .405 \\
\bottomrule
\end{tabular}
}
\caption{Ablation study on the weight of the MSE loss. We keep the weight of the SoftCLIP loss and the discriminative loss as 1.0. All methods are trained and tested using four subjects. Metrics are calculated as the average across four subjects.}
\label{ablation_weight}
\end{table}

We first analyze the robustness of our mutual assistance embedder in Table~\ref{ablation_depth}. It can be inferred from the table that varying the number of depths of our embedder will not influence the decoding performance severely, \textit{e.g.}, using 2 Transformer layers or 4 Transformer layers, indicating the robustness of our method. However, if the model is too heavy, such as using 6 Transformer layers, the dataset may become a limitation because it is difficult to train large models using the NSD dataset that contains only tens of thousands of fMRI-to-image pairs. Besides, if the model is too light-weight, \textit{e.g.}, with only one Transformer layer, the decoding performance will drop. The main reason is that the model is not capable of capturing valid cross-subject commonalities. That's why we encourage the community to develop a stronger model for general brain decoding.

We further analyze the effectiveness of our subject discriminator in Table~\ref{ablation_discriminator}. It can be inferred from the table that the model is sensitive to the designs of discriminators because adversarial training is not stable. A simple subject discriminator will not contribute to the subject-invariant feature extraction but would mislead the training. A complex subject discriminator would be too strong that the extractor can hardly cheat the discriminator. We encourage the community to develop a more proper subject-invariant feature extraction module.
Besides, the stimuli of different subjects in the NSD dataset are not shared, posing significant difficulties for cross-subject feature alignment, because there is no explicit guarantee that features extracted from different fMRI signals are inherently subject-invariant. Therefore, how to design the dataset is still worth exploring.

Finally, we analyze the robustness of our method to the weight of the loss functions in Table~\ref{ablation_weight}.
We fix the weight of the adversarial loss and the SoftCLIP loss, and we vary the value of the weight of the MSE loss. It can be reasoned from the table that a small weight can hardly lead to a satisfying decoding performance. The main reason is that the SoftCLIP loss will learn the relative relationship between subjects, which may mislead the MSE alignment. As the value of the MSE loss is small, a larger weight should be assigned to the MSE loss. That's why we select $\lambda_1 = 1e5$ in our paper.

\end{document}